\documentclass[sigconf]{acmart}


\copyrightyear{2022}
\acmYear{2022}
\setcopyright{acmlicensed}\acmConference[CIKM '22]{Proceedings of the 31st ACM International Conference on Information and Knowledge Management}{October 17--21, 2022}{Atlanta, GA, USA}
\acmBooktitle{Proceedings of the 31st ACM International Conference on Information and Knowledge Management (CIKM '22), October 17--21, 2022, Atlanta, GA, USA}
\acmPrice{15.00}
\acmDOI{10.1145/3511808.3557070}
\acmISBN{978-1-4503-9236-5/22/10}

\usepackage{color}
\usepackage{xcolor}
\usepackage{comment}
\usepackage{algorithm,algpseudocode}
\usepackage{rotating}
\usepackage{multirow}
\usepackage{enumitem}
\usepackage{url}
\usepackage{xspace}
\usepackage{hyperref}
\usepackage{glossaries}
\usepackage{subfigure}
\usepackage{amsmath}
\usepackage{tabularx}  
\usepackage{graphicx}
\usepackage{array}
\usepackage{balance}
\usepackage{multirow}
\usepackage[export]{adjustbox}

\DeclareMathOperator*{\argmax}{argmax}
\newcommand{\eg}{{e.g.},\xspace}
\newcommand{\ie}{{i.e.},\xspace}
\newcommand{\etal}{{et al.},\xspace}
\newcolumntype{C}[1]{>{\centering\let\newline\\\arraybackslash\hspace{0pt}}m{#1}}

\usepackage[colorinlistoftodos,prependcaption]{todonotes}
\usepackage{regexpatch}
\makeatletter
\xpatchcmd{\@todo}{\setkeys{todonotes}{#1}}{\setkeys{todonotes}{inline,#1}}{}{}
\newcommand{\cz}[1]{\textcolor{black}{\textbf{} #1}}
\newcommand{\name}{\textsc{QuickSkill}\xspace}

\makeatother

\begin{document}

\title[QuickSkill: Novice Skill Estimation in Online Multiplayer Games]{QuickSkill: Novice Skill Estimation in Online Multiplayer Games}
\author{Chaoyun Zhang}
\authornote{Both authors contributed equally to this work.}
\affiliation{%
    \institution{Tencent}
    \department{}
    \streetaddress{}
    \city{Shenzhen}
    \postcode{}
    \country{China}
}
\email{hidan.zhang@gmail.com}

\author{Kai Wang}
\authornotemark[1]
\affiliation{%
    \institution{Tencent}
    \department{}
    \streetaddress{}
    \city{Shenzhen}
    \postcode{}
    \country{China}
}
\email{wangjinjie722@gmail.com}


\author{Hao Chen}
\affiliation{%
    \institution{Tencent}
    \department{}
    \streetaddress{}
    \city{Shenzhen}
    \postcode{}
    \country{China}
}
\email{fitzhchen@tencent.com}

\author{Ge Fan}
\affiliation{%
    \institution{Tencent}
    \department{}
    \streetaddress{}
    \city{Shenzhen}
    \postcode{}
    \country{China}
}
\email{gefan@tencent.com}

\author{Yingjie Li}
\affiliation{%
    \institution{Tencent}
    \department{}
    \streetaddress{}
    \city{Shenzhen}
    \postcode{}
    \country{China}
}
\email{wallaceyjli@tencent.com}

\author{Lifang Wu}
\affiliation{%
    \institution{Tencent}
    \department{}
    \streetaddress{}
    \city{Shenzhen}
    \postcode{}
    \country{China}
}
\email{danniewu@tencent.com}

\author{Bingchao Zheng}
\renewcommand{\shortauthors}{Zhang, \etal}
\affiliation{%
    \institution{Tencent}
    \department{}
    \streetaddress{}
    \city{Shenzhen}
    \postcode{}
    \country{China}
}
\email{novazheng@tencent.com}

\begin{abstract}
    Matchmaking systems are vital for creating fair matches in online multiplayer games, which directly affects players' satisfactions and game experience. Most of the matchmaking systems largely rely on precise estimation of players' game skills to construct equitable games. However, the skill rating of a novice is usually inaccurate, as current matchmaking rating algorithms require considerable amount of games for learning the true skill of a new player. Using these unreliable skill scores at early stages for matchmaking usually leads to disparities in terms of team performance, which causes negative game experience. This is known as the ``cold-start'' problem for matchmaking rating algorithms.
    
    To overcome this conundrum, this paper proposes \name, a deep learning based novice skill estimation framework to quickly probe abilities of new players in online multiplayer games. \name extracts sequential performance features from initial few games of a player to predict his/her future skill rating with a dedicated neural network, thus delivering accurate skill estimation at the player's early game stage. By employing \name for matchmaking, game fairness can be dramatically improved in the initial cold-start period. We conduct experiments in a popular mobile multiplayer game in both offline and online scenarios. Results obtained with two real-world \cz{anonymized} gaming datasets demonstrate that proposed \name delivers precise estimation of game skills for novices, leading to significantly lower team skill disparities and better player game experience. To the best of our knowledge, proposed \name is the first framework that tackles the cold-start problem for traditional skill rating algorithms.
\end{abstract}


\maketitle
\renewcommand{\thefootnote}{\fnsymbol{footnote}}
\section{Introduction}
Game matchmaking systems aim at searching comparable teammates and opponents for players to form a fair match, such that all teams in a game have similar abilities. They are massively employed in online multiplayer games \cite{claypool2015surrender, veron2014matchmaking, pramono2018matchmaking}, such as League of Legends. A good matchmaking system avoids disparities as far as possible, since weak teams and players will sense tremendous frustration if they are overwhelmed in a game \cite{veron2014matchmaking}. The quality of matchmaking therefore directly affects players' satisfaction and retention \cite{chen2021matchmaking}, and the life cycle of the game \cite{gong2020optmatch}. 

A matchmaking service makes a decision depending on many criteria, whereas the principal factor is the players' skills scores (a.k.a matchmaking rates).  The matchmaking rate (MMR) quantifies players' game abilities by aggregating their historical competition outcomes or/and performance features into a scalar \cite{ebtekar2021elo}, which is used to compare and rank players' strength at the same game event \cite{graepel2006ranking}.  The most classical and popular MMR algorithms employed in online games is the TrueSkill-family \cite{herbrich2006trueskill}. These methods update a player's MMR after each match given the outcome of the game \cite{herbrich2006trueskill}. In general, the accuracy of the MMR grows with the number of games used for learning. 

\begin{figure}[t]
\centering
\includegraphics[width=\columnwidth]{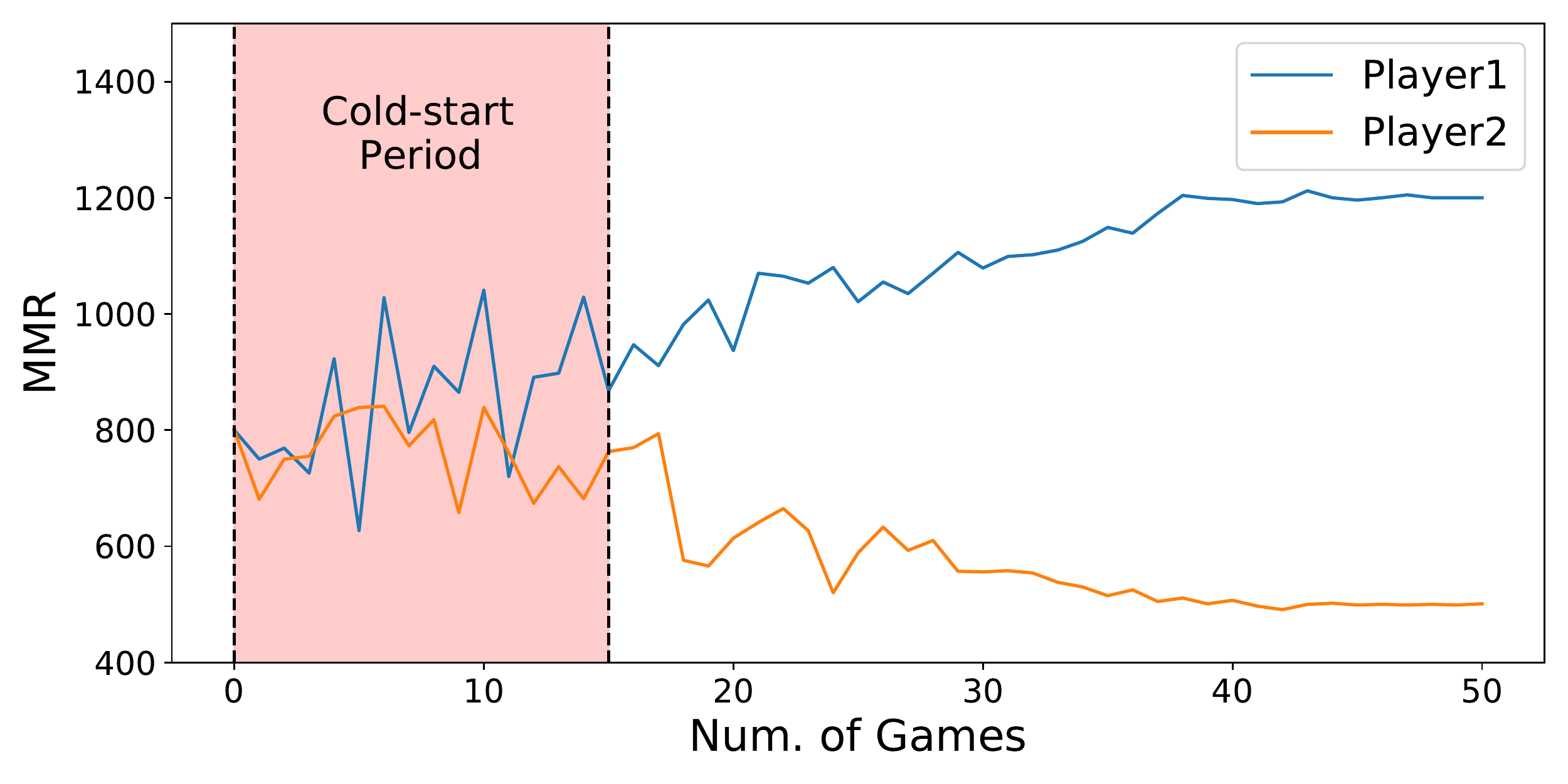}
\vspace*{-1.5em}
\caption{An example of TrueSkill MMRs evolution of two players generated in an online multiplayer game.
\label{fig:illus}}
\vspace*{-1.5em}
\end{figure}
However, industrial applications of TrueSkill suggest that it bears significant shortcomings when assessing skills of novices, especially for online multiplayer games. A new player joins the game with enormous skill uncertainty and limited information, while their game strengths diversify. New players are not always poor players – they may have experience of other similar games. This makes ratings in initial games unreliable, until it observes outcomes of sufficient 
matches. In addition, TrueSkill suffers from unstabitily at the early stage and slow convergence, known as the ``cold-start'' problem \cite{park2019explanatory}\cite{pankiewicz2020warm}\cite{sarkar2021online}. This dramatically affects the accuracy of the skill rating, which indirectly leads to overwhelming games \footnote[2]{Referring to the games where the level of one team is much higher than another one.} and negative game experience.
Fig.~\ref{fig:illus} shows an example of TrueSkill MMRs evolution of two players who have different strengths in an online multiplayer game. Observe that although their MMRs disperse after 40 games, two curves interweave in the cold-start period. This indicates that the two players with huge ability difference may be matched in a game, which causes significant unfairness. 

Estimating the game skill for a new player is not straightforward. First, the strength of a player is summarized via various game performance features. As TrueSkill-based algorithms build upon Bayesian graphical models, working with a large features space will lead to a huge probabilistic graph model. This makes the inference problem mathematically intractable. Therefore, these models are usually oversimplified, which fail to reflect the player's ability in diverse perspectives \cite{deng2021globally}. In addition, a player's skill is less relevant to the side information of the player, such as registration data. Skills in dissimilar games also bring little of help for inferring the player's skill in the targeted domain, as these games may have completely different contexts.
As such, legacy methods that tackle the user cold-start for recommender systems (\eg \cite{wang2018billion,xie2020internal,fan2022field}), are not applicable to the skill estimation. This further complicates the cold-start problem in game skill rating context, making it a ``Achilles' Heel'' for traditional models.

\cz{Though traditional MMR algorithms are unreliable in the initial games, their accuracy grows with the number of matches for learning, and converges after certain games. We observe that converged MMRs estimated by TrueSkill-based models after the cold-start period can better reflect the real game abilities of players at their novice phases (see Sec. \ref{sub:future}). This makes future converged MMRs appropriate labels, which can be learned and correlated with the players' performance in early games using machine learning.}

\textbf{Contributions.} To exploit the future converged MMRs and remedy the cold-start issue, this paper introduces \name, a fast and accurate deep learning based skill estimation framework tailored to new players for online multiplayer games. Unlike TrueSkill and its variants, which only take binary outcomes of the match or limited performance features for learning, \name extracts comprehensive sequential in-game features from each combat match, to profile the multi-dimension game ability of a player. We design a dedicated deep learning model MMR-Net, which accepts feature snapshots collected at different time slices in a game as inputs, and predicts the ``future'' MMR provided by the TrueSkill family of the player. These look-ahead MMRs are more accurate than their counterparts in the cold-start period. By substituting original skill ratings with predicted MMRs, proposed \name delivers more precise skill estimation, which significantly improves the fairness of matchmaking. This buffers the infancy stage of TrueSkill, allowing them to improve during the initial learning period. 

We evaluate our proposed framework on \cz{anonymized} benchmark datasets collected in two different game modes in a well-known online mobile Multiplayer Online Battle Arena (MOBA) game, for both offline and online scenarios. Experiments show that compared to TrueSkill and its variants, our \name achieves more accurate estimation of players' skills for novices in the cold-start period. By employing the proposed framework for matchmaking, over 20\% overwhelming games at novice stages can be eliminated, leading to significantly better game fairness and player game experience. \cz{Importantly, proposed \name can be efficiently deployed as an online machine learning service with limited computing resources requirement.}
To the best of our knowledge, \emph{proposed \name is the first framework that tackles the cold-start problem for traditional MMR models.}

\section{Related Work}
In this section, we review related research and industry practice on game matchmaking and player skill rating.

\subsection{Player Skill Rating}
Most game matchmaking systems build upon accurate player skill rating, which profiles the overall strength of an individual player with a matchmaking rate (MMR).  Classical MMR algorithms are based on Bayesian probabilistic graphical models. In this case, the MMR of each new player is initialized with a unified score, and is gradually updated after each game. Popular MMR algorithms include ELO \cite{bradley1952rank}, ELO-MMR \cite{ebtekar2021elo}, TrueSkill \cite{herbrich2006trueskill}, Glicko \cite{glickman1995glicko} , and their variants or extensions (\eg \cite{minka2018trueskill}\cite{glickman2012example}). 

The well-known ELO system \cite{bradley1952rank} has been widely employed to estimate players' abilities in two-player games, \eg  GO and chess. The TrueSkill \cite{herbrich2006trueskill} model extends Bayesian rating systems to a multiplayer manner, which plays an important role in matchmaking for many different games. These algorithms however, only update with binary win-loss outcomes of matches, while ignoring the performance of players and are biased in many cases. TrueSkill2 \cite{minka2018trueskill} utilizes fine-grained features for the skill estimation, by leveraging  players' performance and correlating around-game features. This delivers more accurate estimation and enables faster convergence.

As aforementioned rating systems require considerable amount of games for learning, the skill estimation for novices are generally inaccurate \cite{park2019explanatory}\cite{pankiewicz2020warm}\cite{sarkar2021online}. Research toward estimating beginners' skills in online games remains largely unexplored.

\subsection{Matchmaking in Online Multiplayer Games}
With players' MMRs provided, the matchmaking system \cite{manweiler2011switchboard} forms a fair game by finding comparable players for each competition with similar total team MMRs. A sophisticated matchmaking system not only considers the MMR scalar of each player in the matchmaking pool, but also the player profiles and team cooperative effects inherent to multiplayer matches (\eg \cite{gong2020optmatch,sapienza2019deep}). This usually leads to more equatable games. For instance, Delalleau \etal employ a neural network to predict player enjoyment to achieve better matchmaking \cite{delalleau2012beyond}. Similarly, research in \cite{gong2020optmatch} completes the matchmaking process at two stages: it first uses a model to predict the outcome of the game, then finds the optimal matchmaking by heuristics sorting.  Deng  \etal \cite{deng2021globally} use reinforcement learning to resolve matchmaking for multiplayer games from the global player pool. This achieves better game equality compared to traditional two-stage approaches.

In general, MMRs remains the core of most matchmaking systems \cite{manweiler2011switchboard}, affecting directly game experience and retention of players \cite{li2021study}. This is particularly important for novices, who have limited number of games to probe their game abilities.

\section{\name: A Dedicated Novice Skill Estimation Framework}
In this section, we first give a brief background and formalize the problem of novice skill estimation.  Next, we introduce the overall architecture of the \name framework. Finally, we present MMR-Nets, the core deep learning predictor of the system, and explain how to learn the game skill of a player with complex, sequential in-game features in a popular mobile MOBA game.

\subsection{Background and Problem Formulation}

\textbf{Multiplayer Online Battle Arena (MOBA) Game}. \cz{MOBA (\eg Defense of the Ancients, Mobile Legends) is one of the most popular game families in both personal computer (PC) and mobile game markets.} It is considered as a mix of real-time strategy, role-playing and action games. In the classical MOBA game setting \cite{mora2018moba}, 5 players controlling different game characters are matched as a team to play against another one. \cz{Players gain coins and become stronger by seizing game resources and slaying their enemies.} All players in a team work collaboratively to destroy the opponent's base as an ultimate goal. \cz{In a MOBA game, skill balance between teams 
has substantial impact on the game experience for players.}

\textbf{MMR and Game Matchmaking}.
The matchmaking rate (MMR) is a score that represents the game skill of a player. Higher MMR means the player are stronger and will generally perform better in the game. Normally, the score is updated after the each match given the outcome of the game, as well as the player's game performance. We formally denote the MMR of the player $p$ at the game $i$ as $s_{p,i}$. This score will be used for  matchmaking at the game $i+1$.

Normally, the MOBA matchmaking system searches two teams comprising multiple players with similar skill levels to construct a combat match. The overall matchmaking rules are complicated, while there are two common criteria should be followed. Namely:
\begin{enumerate}
    \item The players' skills in the same team should be close, \ie $|s_{p, i}^\alpha - s_{p', i'}^\alpha| \leq \tau, \forall p, p'$, and $|s_{p, i}^\beta - s_{p', i'}^\beta| \leq \tau, \forall p, p'$. Here $\alpha$ and $\beta$ are the team indices, $p, p'$ are players in two teams and $\tau$ denotes the player's skill gap threshold.
    \item The summation of skills of all players in a team should be close to another, \ie $|\sum_p s_{p, i}^\alpha - \sum_{p'} s_{p', i'}^\beta| \leq \varphi$, where $\varphi$ is the team's skill gap threshold. Normally, teams with higher skills are supposed to win with higher probabilities, provided that MMRs are accurate.
\end{enumerate}
Matchmaking algorithms used in this paper follow these two basic rules, to guarantee the fairness of a game as much as possible.

\textbf{Key Observation}.
Classical MMR algorithms, \ie TrueSkill \cite{herbrich2006trueskill} and its upgraded variant TrueSkill2 \cite{minka2018trueskill} suffer from the cold-start problem. Their MMRs calculated are relatively inaccurate and unstable before $C$ games. The $C$ is the number of games in the cold-start period and may vary from different MMR algorithms. In this paper, we set $C=18$ for TrueSkill and TrueSkill2 based on our observation on online deployment, as both models converge after 18 games. 

\begin{figure}[t]
\centering
\includegraphics[width=1\columnwidth]{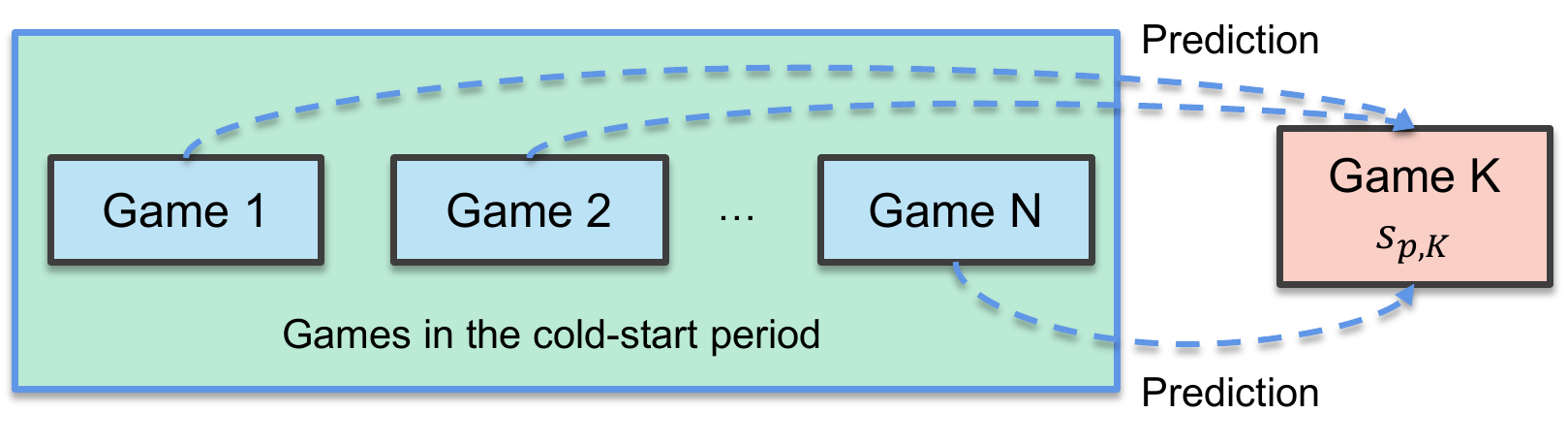}
\caption{An illustration of the objective of \name.
\label{fig:obj}}
\end{figure}

Notably, with the number of games increasing, their MMRs become more precise and stable. We observe that these future MMRs naturally become a satisfying indicators that can better reflect the actual game strength of a player at the novice stage. If we foresee players' future MMRs and use them for  matchmaking in the cold-start period, the fairness of the game can be substantially improved. Therefore, we design \name to ``gaze into 
the future'',  by predicting players' future MMR scores given their game performance at novice stages. The inferred MMRs replace the original TrueSkill rating and are employed for matchmaking at their novices stages.  We illustrate the principle of \name in Fig.~\ref{fig:obj}.

\textbf{Problem Formulation}.
\cz{Predicting the future MMR of a player requires his/her performance feature snapshots at different time in novice games.} Denoting the targeted MMR label in the look-ahead game $K$ for player $p$ as $s_{p, K}$, and the game performance features at time slice $t$ of the game $i$ as $X_{p, i}^t$, we formulate the problem as:
\begin{align}
    \hat{s}_{p, i} = \argmax_{s_{p, K}}\mathsf{p} (s_{p, K} | X_{p, i}^1,\cdots,X_{p, i}^T).
\end{align}
Here, $\mathsf{p}$ is the probability distribution, $\hat{s}_{p, i}$ is the predicted future MMR of at the game $i$ for player $p$, and will be used for matchmaking before game $C$. Note that game $K$ is the future game used for the MMR ground truth, while game $i$ is the current match.
We conduct detailed analysis and experiments of the choice of different future $K$ in Sec.~\ref{sub:future}. The base MMR algorithms employed for the label $s_{p, K}$ are flexible, where we select TrueSkill and TrueSkill2 in this study. 

\subsection{\name in a Nutshell}
\name is a novel deep learning framework that specifically designed for novice game skill rating. It includes two important components, namely \emph{(i)} offline training; and \emph{(ii)} online serving.

\textbf{Offline Training.} In the offline training process, \name collects large amount of game performance data of players at their novice stages, and labels each sample with the players' future MMR $s_{p, K}$. In the case of MOBA games, players' performance is profiled by their statistics at different stages of a game \cite{mora2018moba}. To fully capture the player's performance, we collect game feature snapshots every 3 minutes to construct comprehensive profiles of a player.  Next, we train a dedicated transformer-based MMR-Nets with the aforementioned features by minimizing the mean squared error between the predicted $\hat{s}_{p, i}$ and the ground truth $s_{p, K}$. 

\begin{figure}[t]
\centering
\includegraphics[width=1\columnwidth]{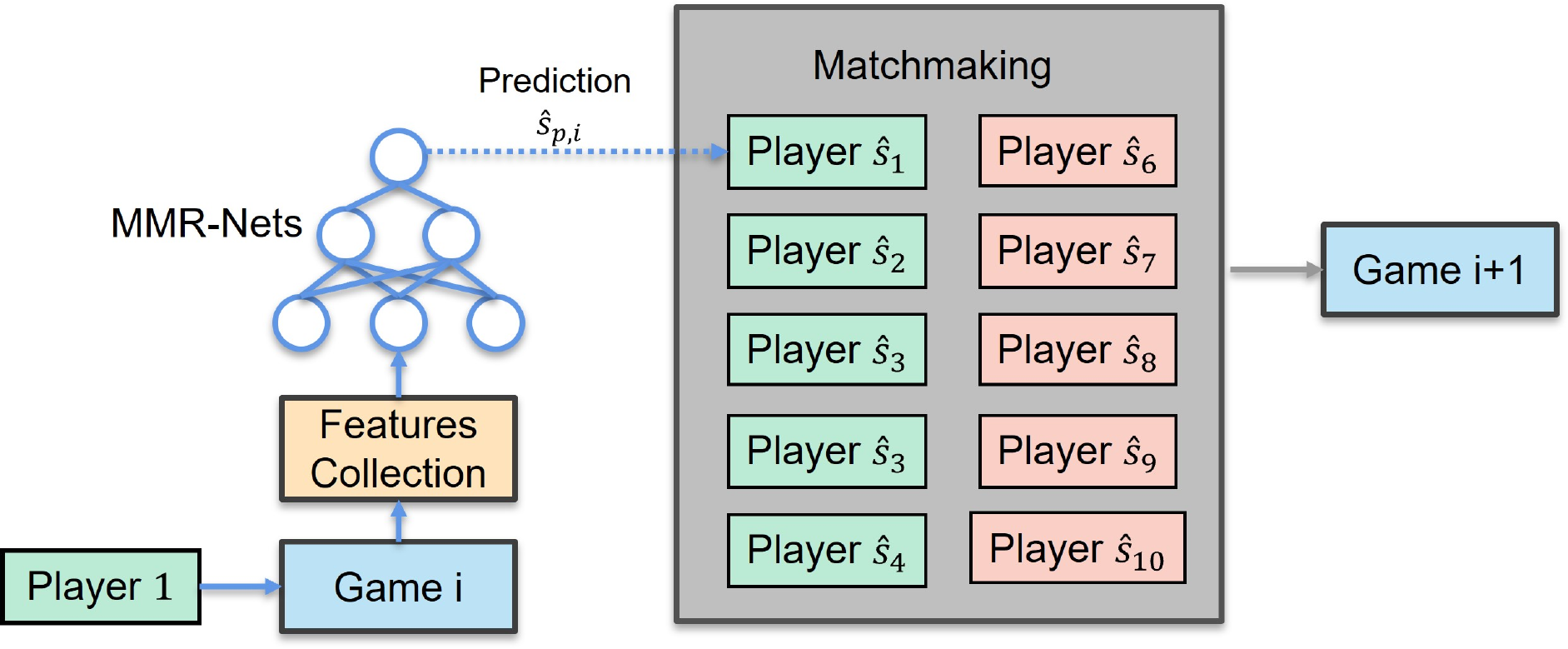}
\caption{An illustration of the online serving process.
\label{fig:online}}
\end{figure}

\textbf{Online Serving.} At the stage of online serving, \name predicts a MMR $\hat{s}_{p, i}$ after each game $i$ in the cold-start period, given the sequential performance features of the players. The $\hat{s}_{p, i}$ substitutes the raw TrueSkill MMR and is employed for the matchmaking phase in the following games $i+1$. We show the online serving process in Fig. \ref{fig:online}. \name buffers the infancy stage of TrueSkill based MMRs models, improving the skill rating at their early learning stages. Once the algorithm that computes the MMR gains sufficient games for learning and moves out of the cold-start period, the matchmaking system switches to the original algorithm for the game matchmaking. This means that \name retires for the corresponding player after game $C$.

\subsection{Feature Design for Skill Learning \label{sec:feature}}
Recall that the objective of a team in a MOBA game play is to destroy their opponent's base. This is however, a complicated process and cannot be accomplished at one stroke.
Strength of one player is quite limited, hence players in a team need to work collaboratively to achieve the ultimate goal \cite{kim2017makes}. These factors make the entire MOBA game play complex.

The skill level of a MOBA player is profiled from different dimensions \cite{deng2021globally}. In this paper, we collect their statistics in a game from various perspectives to fully capture the player's performance and skill. These include \emph{(i)} personal statistics; \emph{(ii)} teammate statistics; and \emph{(iii)} opponent statistics. Personal statistics includes the player's own game-related features, \eg kills, death, gold. These features reflect the player's own performance in a specific game. In addition, performance of teammates and opponents are important references for comparisons between the targeted player and other participants in the same game. We therefore add the teammate and opponent statistics (\eg average team/opponent kills, average team/opponent gold) to the feature set, to distinguish the target with other players. \cz{Note that all features collected in this study do not embrace personal information of players, therefore the data collected does not raise privacy concerns.}

In addition, A MOBA game can be naturally partitioned into various game phases by different time in a game, where each game phase reflects the players' performance in different dimensions \cite{deng2021globally}. For example, at the early phase of the game,  player's statistics represents his skill at the laning stage, while the importance of team battles grows with the game time. The result of final team battle, usually has decisive effects on the game outcome. In order to comprehensively capture the player's game strength, performance at different phases in a game should be leveraged and weighted for model learning. To this end, we sample the aforementioned statistics features snapshot every 3 minutes in a game, to deliver their performance evolution to the MMR-Nets for skill learning, which is detailed next. 

\subsection{Learning Game Skills With MMR-Nets}
\begin{figure}[t]
\centering
\includegraphics[width=0.9\columnwidth]{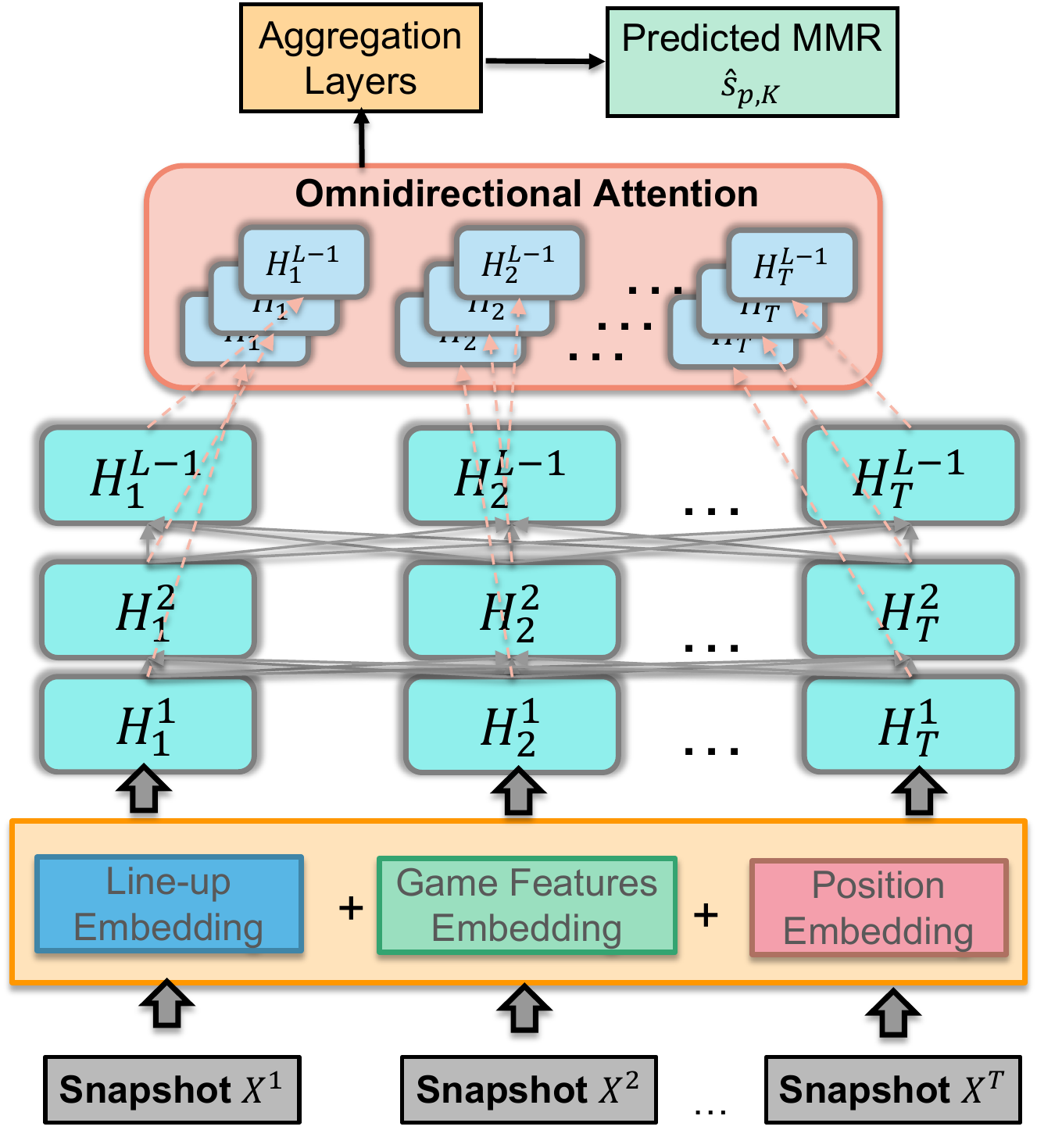}
\caption{The overall architecture of the proposed MMR-Nets.
\label{fig:omninet}}
\vspace*{-1.5em}
\end{figure}

Deep learning has been widely employed to model complex correlations for multivariate time series in different areas (\eg \cite{zhang2020cloudlstm,zhang2019deep,zhang2020microscope,devlin2019bert,zhang2018sequence}). To leverage the heterogeneous sequential game features and achieve accurate skill learning, we design a transformer-based MMR-Net, to extract important information from different feature channels. We show the overall structure of the MMR-Net in Fig.\ref{fig:omninet}.
The proposed MMR-Net includes three important components, namely \emph{(i)} embedding layers; \emph{(ii)} OmniNet layers; and \emph{(iii)} aggregation layers. The model is fed with a sequence of feature snapshots \ie $\{X^1,\cdots,X^T\}$, where each slice includes 115 features, profiling the player's performance from multiple perspectives. 

\textbf{Performance Snapshots Embedding.} The discretized inputs are first processed by an embedding layer, which comprises three components, \cz{namely line-up embedding, game features embedding and position embedding}. The line-up of a MOBA game refers to the character list controlled by each team. Since different characters have synergy or suppression effect on each other \cite{gong2020optmatch}, the line-up  makes significant impact to the player's performance and the outcome of the game \cite{gong2020optmatch}. We therefore isolate the line-up information from other features, and encode them with a dedicated embedding layer. \cz{Game features embedding encodes other sparse performance features of the player in a game into a dense space. The model also includes a regular position embedding, to facilitate the sequential information processing by the transformer structure. Finally, line-up and game features embedding are concatenated and added with the position embedding, which constructs the full embedding for different feature snapshots.}

\textbf{Sequential Learning with OmniNets.} Embedded features are subsequently processed by an OmniNet, which stands for Omnidirectional Representations from Transformers \cite{tay2021omninet}.

\cz{Vanilla transformers build upon multi-head self-attention blocks, which enables to learn contributions of different time slices dynamically. They are widely employed for sequential learning in different applications \cite{li2019enhancing, zerveas2021transformer, zhou2021informer}.  In order to take advantage of this property, we employ the transformer structure as a base to learn the correlation and importance between skills and player's performance of different phases in a game automatically. }

Compared to traditional Transformers \cite{vaswani2017attention}, the OmniNet introduces an additional omnidirectional attention as a meta learner, which connects hidden representations across all layers in the model and weights their importance automatically.
The operation of the omnidirectional attention is formulated as:
\begin{align}
    O_{attention} = \text{Attention}(H_1^1, H_1^2, \cdots, H_N^{L-1}),
\end{align}
where $O_{attention}$ is the output of the omnidirectional attention, and $H_T^{l}$ denotes the hidden layer of the model in layer $l$ at time state $T$. \cz{As the omnidirectional attention operates over all hidden layers of the transformer, the $O_{attention}$ returns a tensor with $(L-1)\times T \times d$ dimensions.}
$O_{attention}$ is added with the output of the final layer of the transformer, \cz{after dimension reduced by a Multiplayer Perception (MLP)}, \ie
\begin{align}
    \text{OmniNet}(X) = \text{Transformer}(X)_{L-1} + MLP(O_{attention}),
\end{align}
where $\text{Transformer}(X)_{L-1}$ is the output of the transformer. The global attention in the OmniNet enables to access the knowledge of the entire network, which helps capture patterns across different features interweaving over different time in a game. This is particularly suitable for the skill learning, as the performance features are inter-correlated across different game phases. Reinforcing such correlations improves the model representability and helps the OmniNet better capture the complex link between performance and game skill. Moreover, the global attention can be regarded as a form of residual learning \cite{huang2017densely}, which is beneficial for the gradient propagation and can be learnt in an end-to-end manner.
 
\textbf{Features Aggregation \& Prediction.}  Finally, outputs from the OmniNet are passed to a dense aggregation layer, to combine the high-level features embedding and make the skill prediction $\hat{s}_{p, i}$. The aggregation layer gathers outputs from all time states, to maximize the sequential information utility. This enables to summarize the player's performance from a global picture of a game. \cz{The entire model is trained by minimising a standard $L2$ loss between inferred skills and ground truth future MMR, \ie }
\begin{align}
    L(\Theta) = \frac{1}{\mathcal{N}} \sum_{\mathcal{N}} (s_{p, K}- \hat{s}_{p, i})^2,
\end{align}
\cz{where $\Theta$ denote the trainable parameters of the MMR-Net, and $\mathcal{N}$ is the number of samples used for training.}

\cz{\textbf{Online Deployment.} Our MMR-Nets are efficient in both offline and online processes. Training the MMR-Net with million-scale data on average requires around 6 hours in a single machine with 8 NVIDIA V100 GPUs. Once trained, the model only needs to be updated every few weeks to meet the online data drift. In the online serving, we deploy 10 containers (4 cores, 2G memory) for feature pre-processing and 15 containers (4 cores, 2G memory) for TF-serving. The average response time of \name is around 20 ms. This is sufficient to meet the requirement of the mobile MOBA game, as updated MMRs are only required in the matchmaking for the next game, which happens after a few seconds from the current game at the soonest.}

\section{Experiments}
To evaluate the performance of \name, we employ datasets collected in a well-known mobile MOBA game for two different game modes.  We provide comprehensive comparisons with different baseline deep learning models, as well as two state-of-the-art MMR algorithms from multiple perspectives. All models are implemented using TensorFlow \cite{tensorflow2015-whitepaper}. We train all architectures with a computing cluster with 8 NVIDIA V100 GPUs.


\subsection{Datasets \label{sub:dataset}}
As current publicly available dataset does not contain sufficient game features required in this study, we collect new heterogeneous datasets in a well-known mobile MOBA game\footnote[3]{Due to the non-disclosure agreement, the name of the game employed for the case study cannot be disclosed.}
for two different game modes, \ie normal and ranking. We show summary statistics of the datasets in Table \ref{tab:stat}. Both datasets are collected in 20 days, where data collected in the first 13 days are employed for training and validation, and the rest are for testing. This generates two large datasets with up to 80 million samples, which are sufficient for the model training and evaluation.

\begin{table}[h]
\centering
\caption{Statistics of the online MOBA dataset. \label{tab:stat}}
\begin{tabular}{c|c|c|c}
\hline
Mode    & Training & Validation & Testing \\ \hline
Normal  & 34M      & 4M         & 14M     \\
Ranking & 44M      & 8M         & 28M     \\ \hline
\end{tabular}
\end{table}

We note that both normal and ranking game modes share the same game context and objectives, while players usually treat two modes with different seriousness.  In the normal game mode, players tend to be more casual and sometimes choose unskilled characters for practice and entertainment. On the contrary, players take the ranking mode more seriously, as it provides more explicit grade to the players to exhibit their game level. For each data, we collect 115 features for each snapshot, where snapshots are sampled every 3 minutes in a game. Each data sample only retains the last 12 feature snapshots, to shear overlength games. Therefore, the dimension of the input fed to the model becomes $115\times12$. MMRs estimated by TrueSkill and  TrueSkill2 after the cold-start periods will be also collected as true labels for the skill learning.

\cz{As a final remark on data collection, we stress that all data collections were carried out in compliance with applicable regulations. In addition, the dataset we employ for our study does not contain personal information about individual players. This implies that the dataset is fully anonymized and desensitized, thus its use for our purposes does not raise privacy concerns.}

\subsection{Benchmarks and Performance Metrics \label{sub:metrics}}
We implement several baseline models for comparison, and design dedicated metrics to comprehensively quantify the performance of proposed \name.

\textbf{Benchmarks.} We compare the performance of our proposed \name with two classical MMR algorithms, namely Trueskill \cite{herbrich2006trueskill} and Trueskill2 \cite{minka2018trueskill}. In particular,
\begin{enumerate}
    \item \textbf{Trueskill} is a probabilistic MMR model that updates a player's rating depending only on the game outcome (win/lose). It is one of the most popular MMR algorithms and widely employed in online multiplayer games.
    \item \textbf{Trueskill2} evolves from its ancestor by introducing extra features into the model, such as player experience and skill in other game modes. It usually converges faster with higher accuracy compared to Trueskill. 
\end{enumerate}
We train proposed \name with labels obtained from both TrueSkill and TrueSkill2 for comparisons.

Deep learning models employed in \name are flexible. We compare the proposed MMR-Net with different machine learning models, namely Linear Regression (LR), Multilayer Perceptron (MLP), Gated Recurrent Unit (GRU) and Transformer. MLP is the most simple neural network architecture that has been widely employed from different purposes \cite{khotanzad1990classification}. GRU \cite{cho2014properties} is a popular model for time series modelling. We use this architecture to model the sequential game performance features of players. Transformer \cite{vaswani2017attention} is a simplified version of  MMR-Net, as it removes the global attention in the model. MLP, GRU and Transformer are equipped with similar embedding layers.  In addition, we train a simplified version of the MMR-Net with features only collected in the last time slice (MMR-Net$_{end}$) for an ablation study.

We show in Table~\ref{tab:model} the detailed configuration along with the number of parameters for each model considered in this study. We employ standard configuration for LR, MLP and GRU. Transformer and MMR-Net share similar configurations to ensure fair comparison. The MMR-Net$_{end}$ only input the last feature snapshot, while the rest of setting is the same as the MMR-Net. Overall, the number of parameters is close for GRU, Transformer and MMR-Net, while the MLP has the most number of parameters.
\begin{table}[t]
\caption{Configuration of all models considered in this paper. \label{tab:model}}
\begin{tabular}{|c|C{4cm}|c|}
\hline
\textbf{Model}  & \textbf{Configuration}                                                                  & \textbf{Parameters} \\ \hline
LR              & A standard LR model with one-hot features.                                              &    1381             \\ \hline
MLP             & Three hidden layers, with (128, 256, 256) hidden units for each layers.                 &    3,659,501          \\ \hline
GRU             & Two hidden layers with (128, 258) hidden units for each layers.                         &    1,926,765          \\ \hline
Transformer     & 6 hidden layers, model dim = 160, heads = 10, with sequence length = 12.                &    1,968,301          \\ \hline
MMR-Net         & 6 hidden layers, model dim = 160, heads = 10, with sequence length = 12.                &    2,071,347          \\ \hline
MMR-Net$_{end}$ & 6 hidden layers, model dim = 160, heads = 10, with sequence length = 1.                 &    1,763,907          \\ \hline
\end{tabular}
\end{table}

\begin{table*}[t]
\caption{The mean±std of MAE, MSE as well as win-rate$_{h}$ of the team with high predicted MMRs (the higher the better) across all models considered, evaluated on mobile MOBA datasets collected in normal and ranking game modes  with $K=18$.\label{tab:res}}
\fontsize{8.5}{7.5}\selectfont
\begin{tabular}{|>{\hspace{-0.3pc}}c<{\hspace{-3.2pt}}|>{\hspace{-0.3pc}}c<{\hspace{-3.2pt}}>{\hspace{-0.3pc}}c<{\hspace{-3.2pt}}>{\hspace{-0.5pc}}c<{\hspace{-3.2pt}}>{\hspace{-0.3pc}}c<{\hspace{-3.2pt}}>{\hspace{-0.3pc}}c<{\hspace{-3.2pt}}>{\hspace{-0.3pc}}c<{\hspace{-3.2pt}}|>{\hspace{-0.3pc}}c<{\hspace{-3.2pt}}>{\hspace{-0.3pc}}c<{\hspace{-3.2pt}}>{\hspace{-0.3pc}}c<{\hspace{-3.2pt}}>{\hspace{-0.3pc}}c<{\hspace{-3.2pt}}>{\hspace{-0.3pc}}c<{\hspace{-3.2pt}}>{\hspace{-0.3pc}}c<{\hspace{-3.2pt}}|}
\hline
\multirow{3}{*}{Model} & \multicolumn{6}{c|}{Normal}                                                                                                                                                                                                                                      & \multicolumn{6}{c|}{Ranking}                                                                             \\ \cline{2-13} 
                       & \multicolumn{3}{c|}{TrueSkill}                                                                                                            & \multicolumn{3}{c|}{TrueSkill2}                                                                                      & \multicolumn{3}{c|}{TrueSkill}                                & \multicolumn{3}{c|}{TrueSkill2}          \\ \cline{2-13} 
                       & MAE                                     & MSE                                     & \multicolumn{1}{c|}{win-rate$_{h}$}                         & MAE                                     & MSE                                     & win-rate$_{h}$                         & MAE           & MSE           & \multicolumn{1}{c|}{win-rate$_{h}$} & MAE           & MSE           & win-rate$_{h}$ \\ \hline
LR                     & 0.61$\pm$0.20                           & 0.40$\pm$0.11                           & \multicolumn{1}{c|}{51.3\%}                           & 1.42$\pm$0.61                           & 2.08$\pm$1.02                           & 50.4\%                           & 1.41$\pm$0.66 & 2.00$\pm$1.03 & \multicolumn{1}{c|}{51.2\%}   & 1.48$\pm$0.70 & 2.22$\pm$1.32 & 52.8\%   \\
MLP                    & 0.56$\pm$0.18                           & 0.33$\pm$0.09                           & \multicolumn{1}{c|}{54.2\%}                           & 1.09$\pm$0.31                           & 1.20$\pm$0.45                           & 60.8\%                           & 0.72$\pm$0.32 & 0.55$\pm$0.21 & \multicolumn{1}{c|}{54.1\%}   & 1.40$\pm$0.56 & 2.01$\pm$1.15 & 60.4\%   \\
GRU                    & 0.54$\pm$0.19                           & 0.30$\pm$0.09                           & \multicolumn{1}{c|}{54.6\%}                           & 1.07$\pm$0.32                           & 1.17$\pm$0.41                           & 61.1\%                           & 0.69$\pm$0.26 & 0.51$\pm$0.18 & \multicolumn{1}{c|}{54.3\%}   & 1.37$\pm$0.52 & 1.90$\pm$1.09 & 61.0\%   \\
Transformer            & 0.51$\pm$0.18                           & 0.28$\pm$0.08                           & \multicolumn{1}{c|}{55.2\%}                           & 1.06$\pm$0.31                           & 1.07$\pm$0.40                           & 61.3\%                           & 0.69$\pm$0.24 & 0.50$\pm$0.18 & \multicolumn{1}{c|}{54.4\%}   & 1.34$\pm$0.50 & 1.86$\pm$1.08 & 61.1\%   \\
MMR-Net                & \textbf{0.49$\pm$0.18} & \textbf{0.25$\pm$0.08} & \multicolumn{1}{c|}{\textbf{55.4\%}} & \textbf{1.02$\pm$0.30} & \textbf{1.04$\pm$0.36} & \textbf{61.6\%} & \textbf{0.66$\pm$0.23} & \textbf{0.47$\pm$0.17} & \multicolumn{1}{c|}{\textbf{54.7\%}}   & \textbf{1.30$\pm$0.50} & \textbf{1.75$\pm$1.06} & \textbf{61.5\%}   \\ \hline
MMR-Net$_{end}$        & 0.59$\pm$0.21                           & 0.37$\pm$0.11                           & \multicolumn{1}{c|}{53.5\%}                           & 1.18$\pm$0.39                           & 1.42$\pm$0.51                           & 59.8\%                           & 0.70$\pm$0.24 & 0.53$\pm$0.20 & \multicolumn{1}{c|}{54.0\%}   & 1.40$\pm$0.53 & 1.96$\pm$1.12 & 60.2\%   \\ \hline
MMR$_{ori}$          & 0.66$\pm$0.27                           & 0.46$\pm$0.14                           & \multicolumn{1}{c|}{51.4\%}                           & 1.23$\pm$0.44                           & 1.56$\pm$0.66                           & 54.0\%                           & 1.41$\pm$0.67 & 2.02$\pm$0.99 & \multicolumn{1}{c|}{51.0\%}   & 1.41$\pm$0.55 & 1.98$\pm$1.12 & 53.6\%   \\ \hline
\end{tabular}
\end{table*}
\textbf{Performance Metrics.} The accuracy of a MMR algorithm can be only quantified by indirect indicators, as it is difficult to measure a player's game skill with an explicit value. To this end, we design different performance metrics to evaluate the performance of proposed \name from three different perspectives, namely \emph{(i)} model precision; \emph{(ii)} game outcome correlation; and \emph{(iii)} effects of players' game experience. We employ different metrics to evaluate the performance of our framework, as detailed next.

Since skill estimation is modelled as a regression problem, we select the classical Mean Absolute Error (MAE) and Mean Square Error (MSE) to assess the precision of the model. MAE and MSE are computed to evaluate the difference between the predicted MMRs $\hat{s}_{p, i}$ and the ground truth $s_{p, K}$.

Normally, teams with higher total MMRs are supposed to win a match with a higher probability, if estimated MMRs for individual player are accurate. The correlations between winning probabilities and team MMR difference therefore become a good indirect indicator that reflectS the precision of estimated MMRs \cite{minka2018trueskill, deng2021globally}. We formally define the win rate
as:
\begin{align}
    \mathrm{Win\ rate} = \frac{\mathbb{N}_{Wins}}{\mathbb{N}} . 
\end{align}
Here $\mathbb{N}_{Wins}$ denotes the number of games where a given team is the winner, and $\mathbb{N}$ is the total number of games used for the evaluation. 

Recall that matchmaking systems serve to improve the players' game experience. Lastly, we evaluate if players' game experience can be improved by substituting traditional MMR algorithms with  \name. This is the most vital metric of matchmaking. In this study, we use the difference between a player's kill and death for this purpose, \ie $|KD| = |\mathbb{K}-\mathbb{D}|$, where $\mathbb{K}$ and $\mathbb{D}$ are the number of kills and deaths of the character controlled by the player. In a MOBA game, a player always try to kill the opponents and avoid death. Therefore, large values of $|KD|$ mean a player is matched with incompatible teammates or opponents, which may lead to negative experience for the player or/and other participants \cite{veron2014matchmaking}.

\subsection{Offline Results on Novice Skill Estimation \label{sub:offline}}
We first train the proposed MMR-Nets and other baselines with MMRs computed by TrueSkill and TrueSkill2 at  game 18, \ie $K=18$, which are the MMRs obtained at the end of the cold-start period ($C=18$). We also investigate how \name behaves when training with different MMR labels, \ie TrueSkill and TrueSkill2, as well as the influence of training with sequential features. We report the mean and standard deviation of MAE and MSE, and the win rates of the teams with higher predicted MMRs of a game (denoted as win-rate$_{h}$) on both datasets in Table~\ref{tab:res}. The precision of the MMRs model has positive correlations with the win-rate$_{h}$ metric used in the table. MMR-Net$_{end}$s are only trained with end-game feature snapshots, and MMR$_{ori}$ denotes original MMR values estimated by TrueSkill/TrueSkill2 in each game.
Observe that the proposed MMR-Nets in general obtain superior performance over other benchmark models for both datasets, as they achieve lower MAE/MSE and higher win-rate$_{h}$. This suggests that the omnidirectional attentions improve the representability of the model, which enables better encoding of the heterogeneous game features. In addition, sequential models, including MMR-Net, Transformer and GRU, achieve better performance over the MLP and LR. In comparison with the MMR-Net$_{end}$, which only employs the last feature snapshot for training, the MMR-Net performs significantly better. This indicates that the skill of a player can be profiled more precisely by evaluating his performance at different phases of the game, demonstrating the superiority of the feature design described in section~\ref{sec:feature}. 

Taking a closer look at Table~\ref{tab:res}, it appears that the original MMR$_{ori}$ estimated after each game achieves the worst performance, which indicates that TrueSkill and TrueSkill2 in the cold-start period are inaccurate and unreliable. As a comparison, our MMR-Nets obtain up to 4\% and 7.9\% higher win-rate$_{h}$, yielding a remarkable improvement.  In addition, MMRs estimated by TrueSkill2 are more accurate than its TrueSkill, as models trained with TrueSkill2 labels achieve significantly better performance in terms of win-rate$_{h}$. In conclusion, by combining the MMR-Nets with TrueSkill2 labels, proposed \name obtains the best results across all metrics on both datasets, by achieving up to 21.1\% lower MAE and 11.2\% higher win-rate$_{h}$ than other baselines. Furthermore, incorporating sequential features at different phases in a game significantly improves the performance of \name.

\begin{figure}[t]
\centering
\includegraphics[width=0.95\columnwidth]{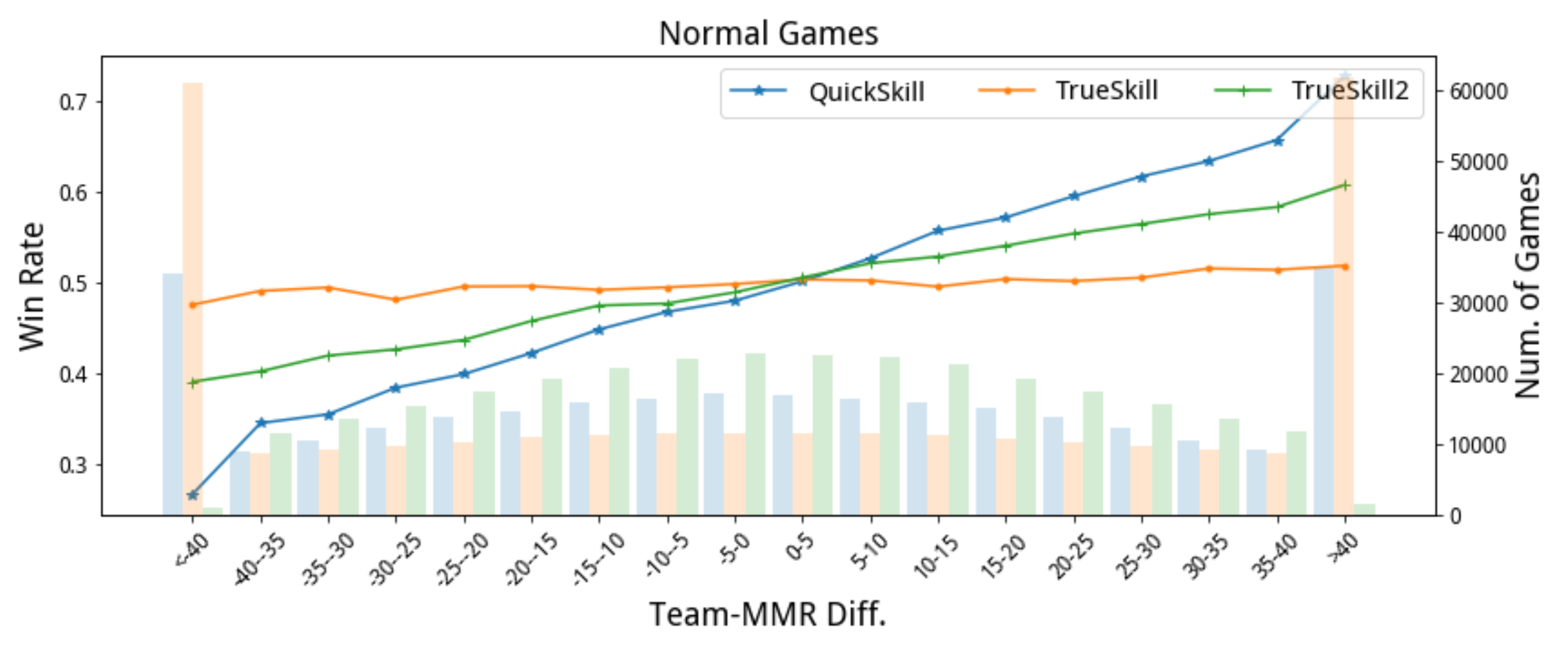}
\vspace*{-1em}
\caption{Team win rates w.r.t. team MMR difference of three MMR models evaluated in the normal mode dataset. The bars in the figures represent the number of games in each bucket.
\label{fig:normal_wr}}
\vspace*{-1.5em}
\end{figure}

\textbf{Correlation of Team Win Rate and MMR Difference.} Next, we investigate the correlation between game outcomes and team MMR differences. As the team MMR is the summation of all players' MMRs in the team, the more accurate the MMR model is, teams with higher total MMRs should win with higher probabilities. This means that win rates should have a stronger positive correlation with the team MMR difference. We compute the win rate of the team as a function of the team MMR difference of three MMR models considered for the normal game mode, as shown in Fig.~\ref{fig:normal_wr}. Observe that team win rates grow with the MMR team difference for all MMR models, as expected. Compared to other baselines, proposed \name exhibits the strongest positive correlation between the MMR difference and win rates, which suggests that \name is more accurate than its counterparts.

Team skill difference generally grows with the team MMR difference $\varphi$. A large $\varphi$ means that the MMR model believes the matchmaking is unfair, and should be avoided. We denote games with team MMR difference $\varphi > 40$ (suggested by the game designer) as ``unfair'' games identified by the MMR models (the leftest and rightest bars in the figure).  Interestingly, TrueSkill senses the most number of unfair games, whereas these games are not truly unfair. This is reflected by the fact that both underdog (lower MMR) and favorite (higher MMR) teams identified by TrueSkill only achieve almost 50\% win rates. TrueSkill2 only senses 0.85\% unfair games, while these games have more extreme win rates. Notably, the proportion of unfair games identified by proposed \name is 20.87\%, and the win rates of these games are the most extreme. This means those games identified by \name were indeed unfair games. By setting the team MMR gap threshold $\varphi=40$, games with $\varphi>40$ will be eliminated, if using our \name for matchmaking. This reduces significant amount of overwhelming games.

\begin{table*}[t]
\caption{Comparisons of win-rate$_{h}$ of the team with high predicted/ground truth MMRs between different $K$ configurations. \label{tab:k}}
\begin{tabular}{|c|cccc|cccc|}
\hline
\multirow{3}{*}{$K$} & \multicolumn{4}{c|}{Normal}                                                      & \multicolumn{4}{c|}{Ranking}                                                     \\ \cline{2-9} 
                     & \multicolumn{2}{c|}{TrueSkill}                 & \multicolumn{2}{c|}{TrueSkill2} & \multicolumn{2}{c|}{TrueSkill}                 & \multicolumn{2}{c|}{TrueSkill2} \\ \cline{2-9} 
                     & \name & \multicolumn{1}{c|}{Ground truth} & \name    & Ground truth    & \name & \multicolumn{1}{c|}{Ground truth} & \name    & Ground truth    \\ \hline
12                   & 54.1\%     & \multicolumn{1}{c|}{58.1\%}       & 60.2\%        & 71.9\%          & 53.9\%     & \multicolumn{1}{c|}{57.5\%}       & 60.2\%        & 72.3\%          \\
15                   & 54.5\%     & \multicolumn{1}{c|}{59.3\%}       & 60.8\%        & \textbf{72.8\%}          & 53.7\%     & \multicolumn{1}{c|}{58.2\%}       & 60.8\%        & 73.8\%          \\
18                   & \textbf{55.4\%}     & \multicolumn{1}{c|}{\textbf{59.6\%}}       & \textbf{61.6\%}        & 72.3\%          & \textbf{54.0\%}     & \multicolumn{1}{c|}{\textbf{58.5\%}}       & 61.5\%        & \textbf{74.0\%}          \\
21                   & 55.3\%     & \multicolumn{1}{c|}{59.4\%}       & 61.3\%        & 70.6\%          & 53.9\%     & \multicolumn{1}{c|}{58.2\%}       & \textbf{61.7\%}        & 72.6\%          \\ \hline

\end{tabular}
\end{table*}

\vspace*{-0.5em}
\subsection{Learning with Different ``Future''\label{sub:future}}
Selecting an appropriate $K$ as targets is not straightforward, as MMR labels $s_{p, K}$ with a small $K$ are inaccurate and unstable, while configuring a large $K$ may cause the labels to over deviate from the players' skills in early games, as players make progress and gain experience in every game. In order to find proper MMRs for learning targets, we tune label $s_{p, K}$ with different $K$, to evaluate the effect of learning with different ``future''. We show the win-rate$_{h}$ results of predictions of \name and corresponding ground truths for both TrueSkill and TrueSkill2 in two datasets in Table~\ref{tab:k}.

Observe that $K=18$ in general achieves the highest win-rate$_{h}$ for both \name and ground truths for most of the cases. The win-rate$_{h}$ grows with the $K$ when $K\leq C$, but become stable afterward. This suggests that simply setting $K=C$ delivers a good performance for the objective of novice skills estimation, and can be employed as a standard configuration. On the other hand, TrueSkill2 always achieves better results over TrueSkill, for both \name prediction and ground truth. This reconfirms that TrueSkill2 provides more accurate and reliable MMR labels for learning.

\vspace*{-0.75em}
\subsection{Online Effect on Players' Game Experience\label{sub:online}}
\begin{figure}[t]
\centering
\includegraphics[width=0.95\columnwidth]{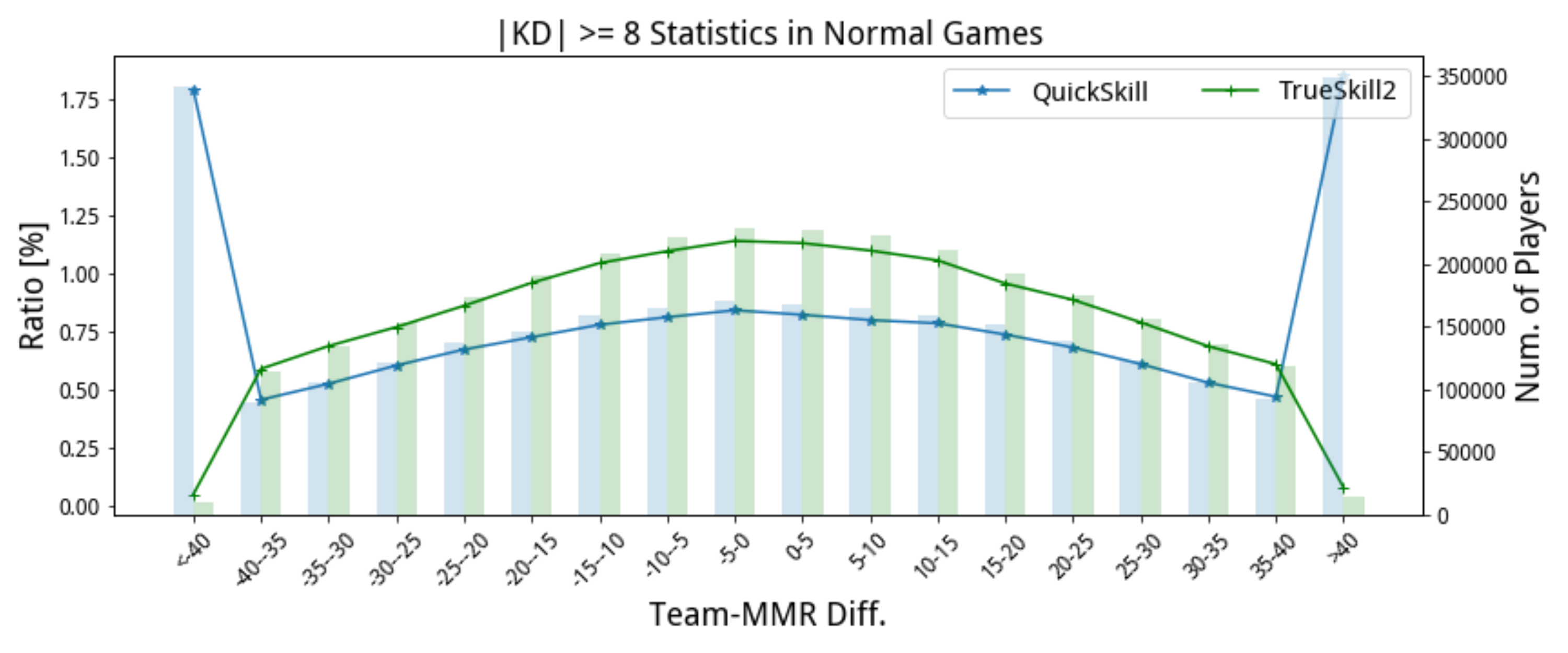}
\vspace*{-1.5em}
\caption{Ratios of players with $|KD|\geq 8$ in a game w.r.t. team MMR difference comparison in the normal mode dataset. Line charts refer to the ratio of players with $|KD|\geq 8$ in each bucket and bar charts represent the number of players.
\label{fig:kd8r_normal}}
\vspace*{-1.5em}
\end{figure}

In a real online game, a player's game experience is dramatically affected by the player's $|KD|=|\mathbb{K}-\mathbb{D}|$. Large values of $|KD|$ suggest that the player overwhelms or is overwhelmed by opponents, leading to negative game experience. MMR algorithms used for matchmaking affect the skills of players, which naturally influence the $|KD|$ of participants. To evaluate the relation between different MMR algorithms employed and players' game experience, we calculate the ratio of players  with $|KD|\geq 8$ as a function of team MMR difference in normal games, as shown in Fig.~\ref{fig:kd8r_normal}. \textbf{These games with extreme $|KD|>8$ result in frustration for new players based on our surveys, and are one of the key reasons for churn \cite{b2c81eac5eaa427ab58be9d7e893a575}.} We select \name trained with TrueSkill2 for a case study.

Similarly, we denote games with team difference $\varphi > 40$ as ``unfair'' games sensed by MMR models. Observe that players with extreme $|KD|$ were quite evenly distributed across difference intervals calculated by MMR TrueSkill2 (green bars), while they are concentrated in \name's intervals at the two poles (blue bars). Specifically, in unfair games identified by \name, the proportion of players with $|KD|\geq 8$ is 3.73\%, while the same metric is 0.08\% for TrueSkill2. This suggests that by setting the team MMR gap threshold $\varphi=40$, using \name for matchmaking can effectively reduce extreme player experience, which significantly improves their satisfaction.

\begin{figure}[t]
\centering
\includegraphics[width=0.95\columnwidth]{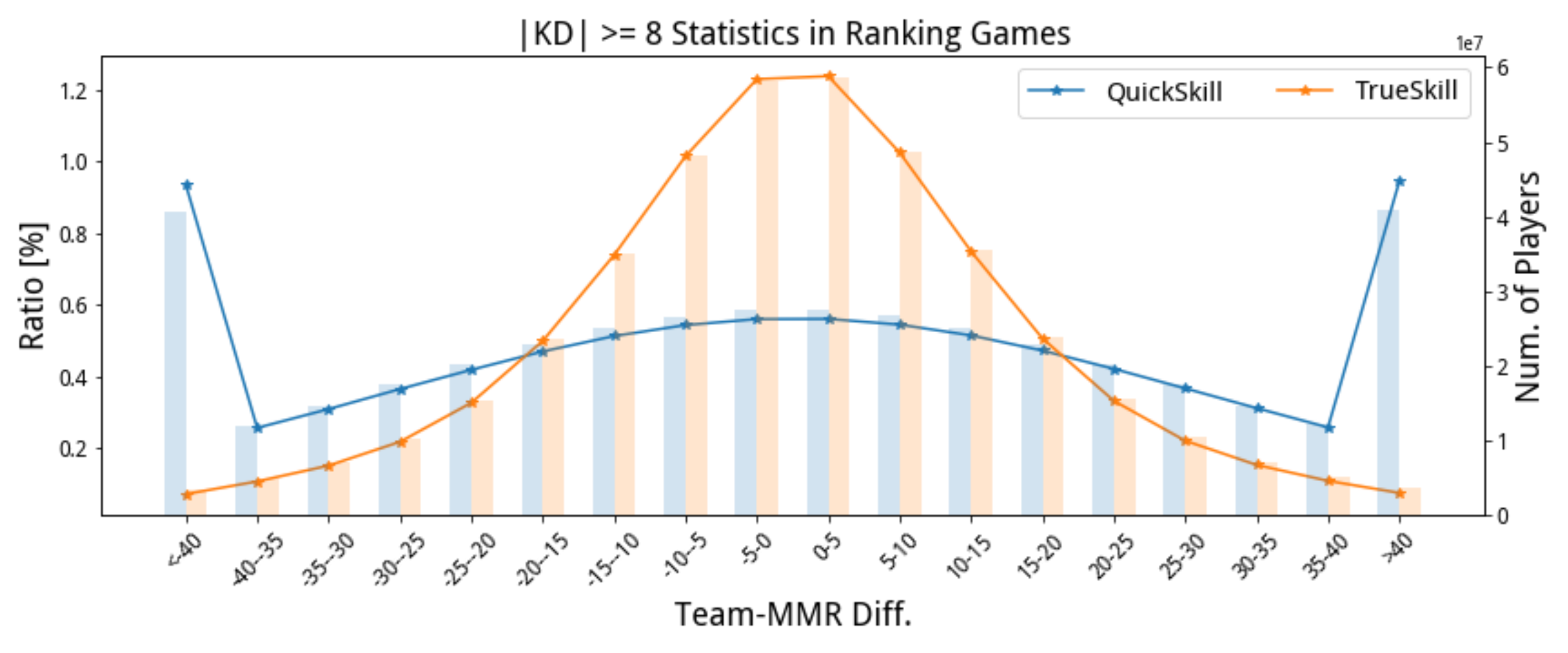}
\vspace*{-1.5em}
\caption{Ratios of players with $|KD|\geq 8$ in a game w.r.t. team MMR difference comparison in the ranking mode dataset. Line charts refer to the ratio of players with $|KD|\geq 8$ in each bucket and bar charts represent the quantity of players.
\label{fig:kd8r_rank}}
\vspace*{-1.5em}
\end{figure}

\begin{figure*}[t]
\centering
\includegraphics[width=1.92\columnwidth]{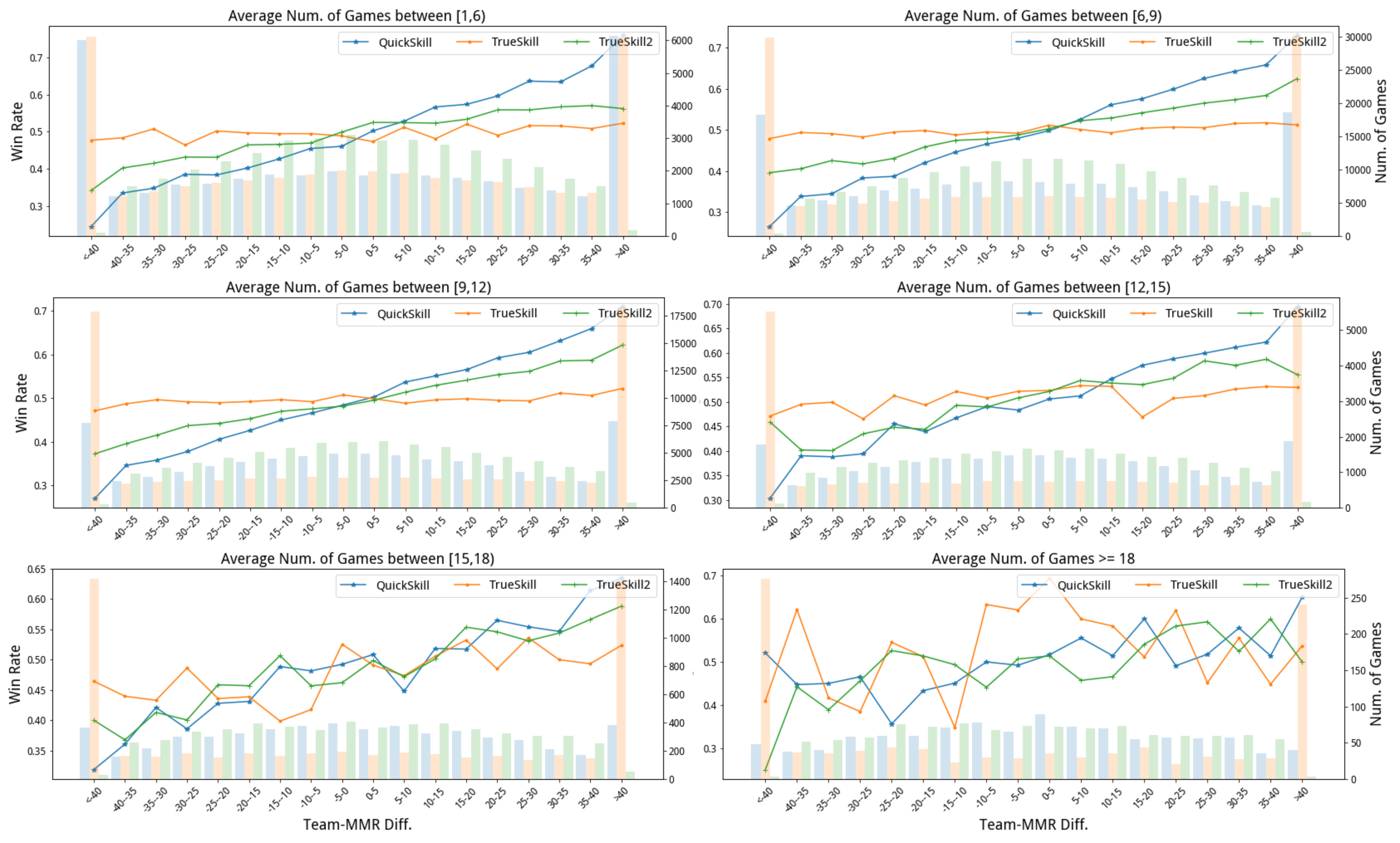}
\caption{Win rate comparison with respect to team MMR difference of three MMR models across different average numbers of games evaluated in the normal mode dataset. Bars in the figures represent the number of games in each difference interval.
\label{fig:mul_normal_kd}}
\end{figure*}


Tuning attention to Fig.~\ref{fig:kd8r_rank}, where we show the ratio of players with $|KD|\geq 8$ as a function of the team MMR difference ranking games. Compared to TrueSkill, the performance of our \name is also remarkable. In the unfair games identified by \name, the proportion of players with $|KD|\geq 8$ is 2.01\%, while the same metric is 0.09\% for TrueSkill. These evaluations prove the effectiveness of \name in a different game mode, which further demonstrates the robustness of the proposed system.

\vspace*{-0.5em}
\subsection{Win Rate Comparison in Cold-start Periods\label{sub:kd_c}}
We now delve deeper into the performance of \name at different stages of the cold-start period. To this end, in Fig.~\ref{fig:mul_normal_kd} we show the correlation between team win rate and MMR difference, where each subplot in the figure collects all games with different average numbers of matches that participants have played. This corresponds to different stages of the cold-start period.  Recall that the team win rate should be  positively related to the team MMR difference, if MMRs calculated for each player are accurate. 

Observe that our \name performs significantly better than other baselines when average game numbers are small (\ie average game number $<15$), while its advantage fades when its counterparts gain more games for learning.
This is reflected by the slope of each curve -- \name exhibits  much stronger correlations between team win rate and MMR difference at early games than other baselines. Their performance becomes fairly close when  average numbers of games is between $[15, 18)$. Interestingly, 18 is exactly the number of games of the cold-start period. After game 18, the system switches to TrueSkill2 for matchmaking. As when exceeding the cold-start period, \name shows no advantages over TrueSkill2, while it has already fulfilled its goal for beginners.


\subsection{Attention Visualization \label{sub:case}}
\begin{figure}[t]
\includegraphics[width=1.05\columnwidth]{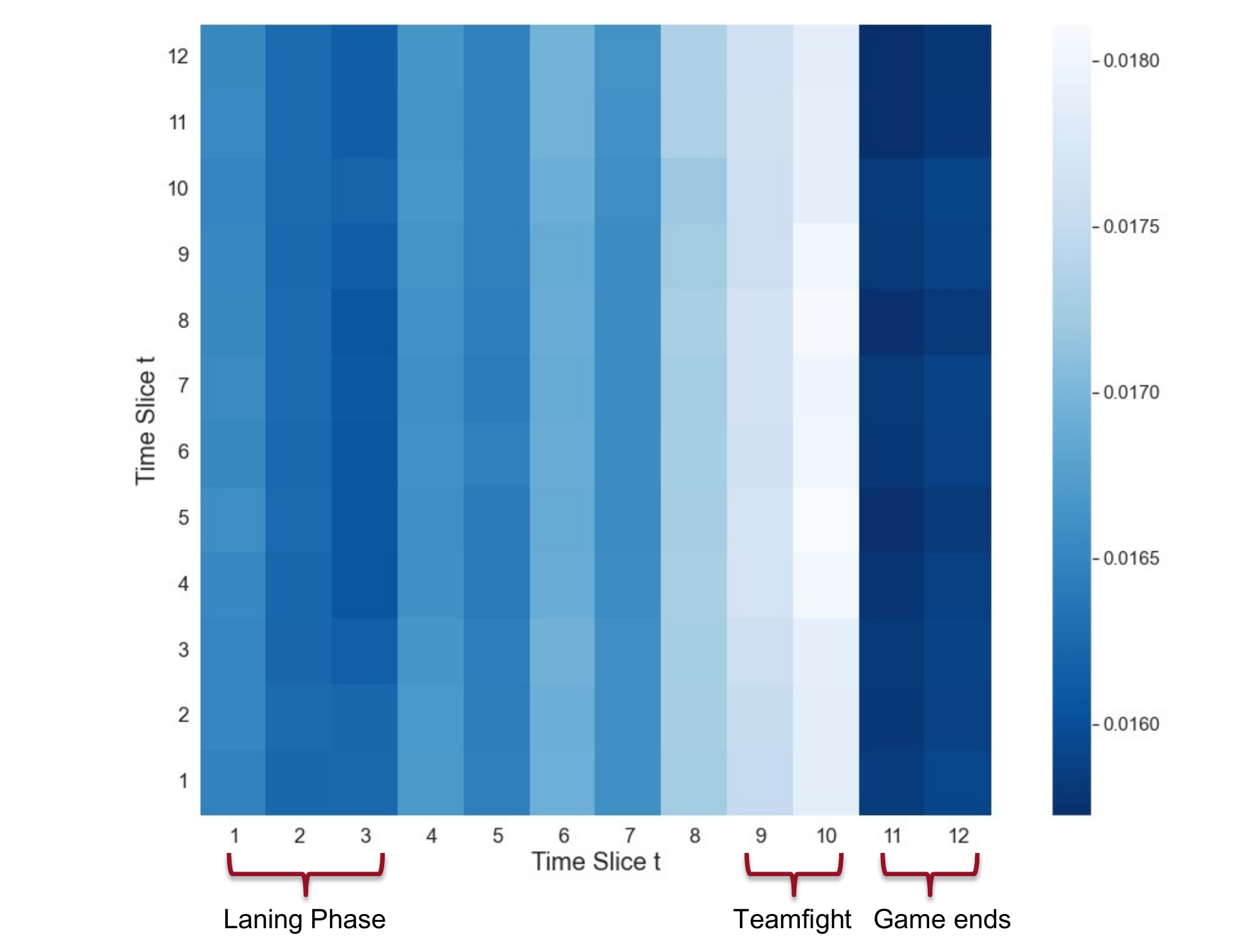}
\caption{Heatmap visualization of the omnidirectional attention module. Weights are averaged over layers.
\label{fig:attention}}
\vspace*{-1.5em}
\end{figure}
We conclude our experiments by examining the exact sequential importance learned by the proposed MMR-Net. To this end, we visualize as a heatmap the averaged weights of the omnidirectional attention module of a sample in Fig.~\ref{fig:attention}. Lighter color represents greater attention weight values. Observe that the MMR-Net commits less attention at the early stage of the game corresponding to the laning phase. The importance grows with time, reaching a peak at the final team battle, which becomes a showdown of the game. The player's performance at these stages highly affects the judgment of the model. The significance drops rapidly with the end of the game, as the outcome becomes a foregone conclusion. This shows that the MMR-Nets indeed capture keys that affect the skill of a player in a MOBA game, which highly meets our expectations.

\section{Conclusions}
This paper proposes \name, an original deep learning based novice skill estimation framework to improve the player skill rating in the cold-start period for traditional MMR algorithms. \name collects multiple feature snapshots in a game to comprehensively profile the performance of players, and predicts their future MMRs, which are generally more accurate and reliable. To model the correlations between players' game abilities and their in-game performance, we design a dedicated model MMR-Net, which evolves the classical Transformer with an omnidirectional attention operator. This augments the representability of the model and enables more accurate MMR predictions. We evaluate the proposed framework on \cz{anonymized} datasets collected in two different game modes of a popular mobile MOBA game. Offline experiment results suggest that \name achieves superior performance over TrueSkill, TrueSkill2 and other baseline models in terms of accuracy. Importantly, our online evaluations show that using \name for matchmaking leads to significantly better game fairness by reducing overwhelming games and  players' extreme game experience. To the best of our knowledge, our \name is the first deep learning based framework that tackles the cold-start problem of skill rating in online multiplayer games.



\newpage
\bibliographystyle{unsrt}
\bibliography{ref}

\end{document}